\DeclareUnicodeCharacter{0301}{*************************************}
\documentclass[10pt,twocolumn,letterpaper]{article}

\usepackage{wacv}              %

\usepackage{graphicx}
\usepackage{amsmath}
\usepackage{amssymb}
\usepackage{booktabs}
\usepackage[accsupp]{axessibility}

\usepackage{color}

\usepackage[pagebackref,breaklinks,colorlinks]{hyperref}
\usepackage{bm}
\usepackage{comment}

\usepackage[capitalize]{cleveref}
\crefname{section}{Sec.}{Secs.}
\Crefname{section}{Section}{Sections}
\Crefname{table}{Table}{Tables}
\crefname{table}{Tab.}{Tabs.}

\usepackage[accsupp]{axessibility}

\def\wacvPaperID{2203} %
\def\confName{WACV}
\def\confYear{2025}

\begin{document}

\title{Dense Depth from Event Focal Stack}

\author{Kenta Horikawa$^{1}$~~~~~Mariko Isogawa$^{1}$~~~~~Hideo Saito$^{1}$~~~~~Shohei Mori$^{2,1}$\\
$^{1}$ Keio University~~~~~$^{2}$ University of Stuttgart\\
{\tt\small hrkenta20008@keio.jp}~~~{\tt\small mariko.isogawa@keio.jp}~~~{\tt\small hs@keio.jp}~~~{\tt\small s.mori.jp@ieee.org}
}
\maketitle

\begin{figure*}[ht]
    \centering
    \setlength\abovecaptionskip{0pt}
    \includegraphics[width=0.95\hsize]{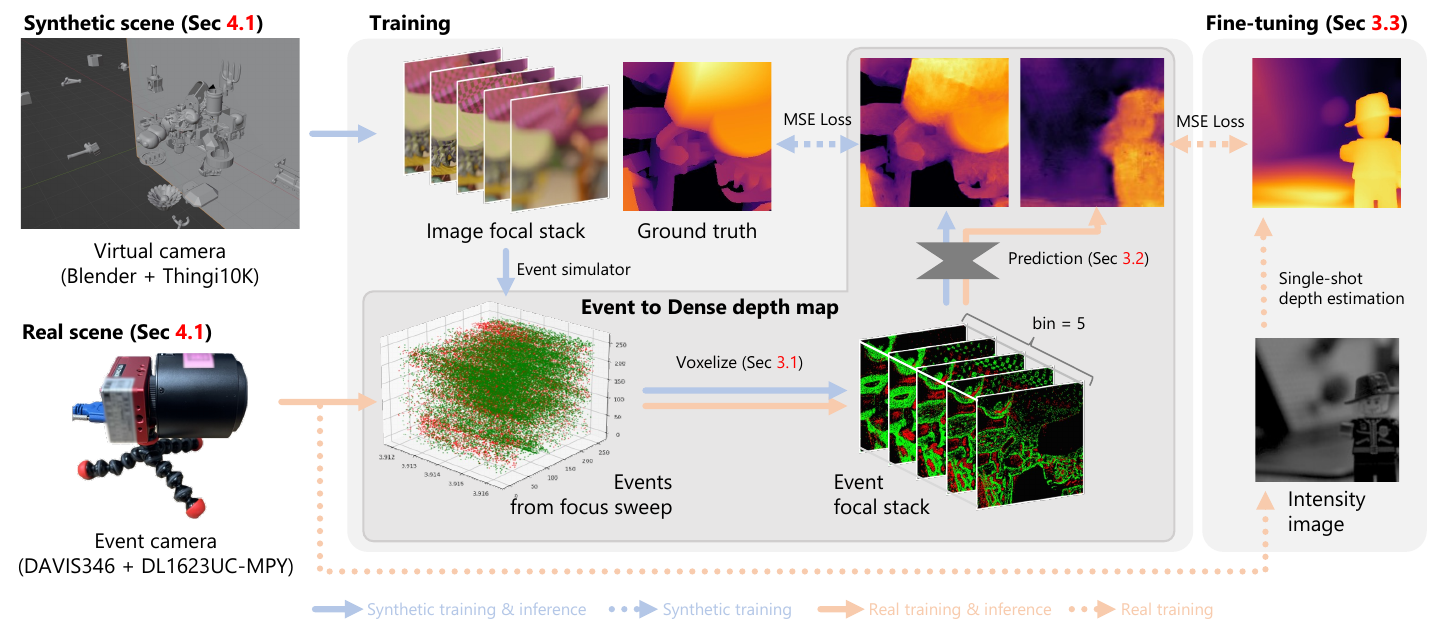}
    \caption{\textbf{The proposed framework for a dense depth map only from an event focal stack.} We collected the datasets in both synthetic and real-world (Sec~\ref{sec:dataset}) environments for this framework. By voxelizing the events from the focus sweep into an event focal stack (Sec~\ref{sec:voxelize}), the data is transformed into a format compatible with a U-Net like CNN architecture and then input into the network (Sec~\ref{sec:network}). We aim to bridge the domain gap between synthetic and real-world data by fine-tuning the model, initially trained on the synthetic dataset, with real-world data (Sec~\ref{sec:finetune}).}
    \label{fig:flow}
\end{figure*}
\setlength\textfloatsep{0pt}

\begin{abstract}
We propose a method for dense depth estimation from an event stream generated when sweeping the focal plane of the driving lens attached to an event camera.
In this method, a depth map is inferred from an ``event focal stack'' composed of the event stream using a convolutional neural network trained with synthesized event focal stacks.
The synthesized event stream is created from a focal stack generated by Blender for any arbitrary 3D scene. This allows for training on scenes with diverse structures. Additionally, we explored methods to eliminate the domain gap between real event streams and synthetic event streams.
Our method demonstrates superior performance over a depth-from-defocus method in the image domain on synthetic and real datasets.
\end{abstract}

\section{Introduction}
\label{sec:intro}
Depth estimation enables various applications, including collision avoidance in autonomous vehicles \cite{m4depth} and 3D interaction in mixed reality \cite{mixed_reality}.
Approaches to this fundamental task are often phrased as ``\textit{depth from X},'' where X can be substituted by stereo \cite{stereo_depth}, defocus \cite{depth_from_defocus}, and even sounds \cite{acoustic_depth}.
One of the most feasible solutions is to use a monocular camera to analyze active motions or disparities from one frame to another.
However, this approach easily fails in extreme conditions such as fast motions and over and under exposures \cite{motion_blur_depth}.
As mobile measurement devices, conventional cameras yet consume electrical power and memory resources too much, limiting further practical applications.

Event-based cameras, or event cameras, collect temporally and spatially occasional responses (i.e., events) and, thus, can break some of the limitations of the cumulative nature of conventional cameras \cite{event_camera}.
They report the positions, time, and polarity of intensity change at an extremely high frequency.
This allows the device to report tens of millions of high-dynamic range responses with significantly less electricity.
Consequently, depth from events can lead to an efficient and robust depth estimation \cite{e2depth,haessig2019spiking,jiang2024learning}.

Therefore, inspired by the previous framework that utilizes RGB image focal stacks, a series of images captured at different focus distances, to infer depth and create all-in-focus images~\cite{image_focal_depth_estimation}, we propose a method for dense depth estimation using an ``event focal stack,'' which consists of a sparse set of events corresponding to in- and out-of-focus regions, driven by active lens control.
Contrary to the existing approaches that use events from optical flows~\cite{e2depth} or estimates depths at exact event locations \cite{haessig2019spiking,jiang2024learning}, we design a network that derives a complete depth map.

To tackle this task, this paper implements an event focal stack simulator for arbitrary scenes using 3D computer graphics software, and event simulators~\cite{simulator_esim, simulator_dvs}. One straightforward approach is to train solely on synthetic data generated by this simulator. However, our analysis revealed a significant domain gap between the events generated by the simulator and real-world events. To mitigate this gap, we introduce fine-tuning using real data.

In summary, we make the following contributions: 
(1) We propose a deep dense depth estimation method using an event focal stack from lens defocus.
To this end, (2) we propose a framework that directly maps an event focal stack to a dense inverse depth image.
For this framework, (3) we created our own synthetic and real-world datasets, consisting of paired event focal stacks and inverse depth images, to perform training and inference.
(4) We also extensively investigated the domain gap between synthetic events from existing simulators and physical events, revealing the limitations of the simulators in our task.
Based on the findings, we introduced a fine-tuning technique to effectively bridge this domain gap, thereby enhancing the model applicability to real-world scenarios. 

\section{Related Work}
\label{sec:related}
This section reviews image- and event-based depth estimation methods, event simulators, and their challenges.

\subsection{Depth from Defocus}
Depth from Defocus~(DfD), pioneered by Pentland ~\cite{depth_from_defocus} is a technique for estimating the scene depth by analyzing the amount of blur (defocus) in images. When an image is captured, objects at different distances appear with varying degrees of sharpness or blur. DfD leverages this information, typically using multiple images with different focus settings, to compute the distance of objects from the camera. 
However, model-based DfD is susceptible to changes in lighting conditions and the scattering of light on surfaces.

Recent data-driven approaches resolve this issue using a deep neural network trained on a synthetic dataset since defocus analysis is independent of the image domain \cite{image_focal_depth_estimation}.
The Focus on Defocus framework introduces image focal stacks, a series of images captured at different focus distances, to infer depth and all-in-focus images \cite{image_focal_depth_estimation}.
A typical U-Net can also learn soft 3D reconstruction , a method that can flexibly perform 3D reconstruction even with missing or noisy data like event, from synthetic focal stacks \cite{Ishikawa2023FS2MPI}.
However, networks trained in the color domain suffer from extreme conditions such as under and over-exposures and motion blur.
Although networks can be trained on images that simulate extreme conditions, optical behaviors are often difficult to simulate appropriately \cite{cv_book}.

Note that recent approaches are powerful enough to estimate a depth map from a single-shot image \cite{midas, first_depth, deep_depth_from_focus}.
Nonetheless, all these approaches rely on the color domain and are weak under extreme conditions \cite{cv_book}.

\subsection{Depth from Events}
Depth-from-event approaches take advantage of the event nature, such as high temporal resolution, high dynamic range response, and low power consumption. Existing works have explored events from motion and used recurrent networks to handle temporal events\cite{e2depth, event_depth_EReformer}.
The idea of \textit{depth from event focal stack} is new, and only a few attempts exist.
These approaches estimate a single focus distance for auto-focus \cite{event_all_in_focus} and a sparse depth \cite{jiang2024learning}\footnote{This arXiv preprint validates their approach only in simulated event data. Instead, we recorded real event focal stacks and used them for fine-tuning and testing.}.

Instead, we estimate a complete depth map from sparsely observed events.
We collect all events from a lens focus sweep that travels across the volume of interest and voxelize them as an event focal stack.
Inspired by the image from events \cite{reconstruction}, we train our network to infer a dense depth map from the stack instead of an intensity image.

\begin{figure*}[tb]
    \setlength\abovecaptionskip{0pt}
    \begin{center}
    \includegraphics[width=0.8\hsize]{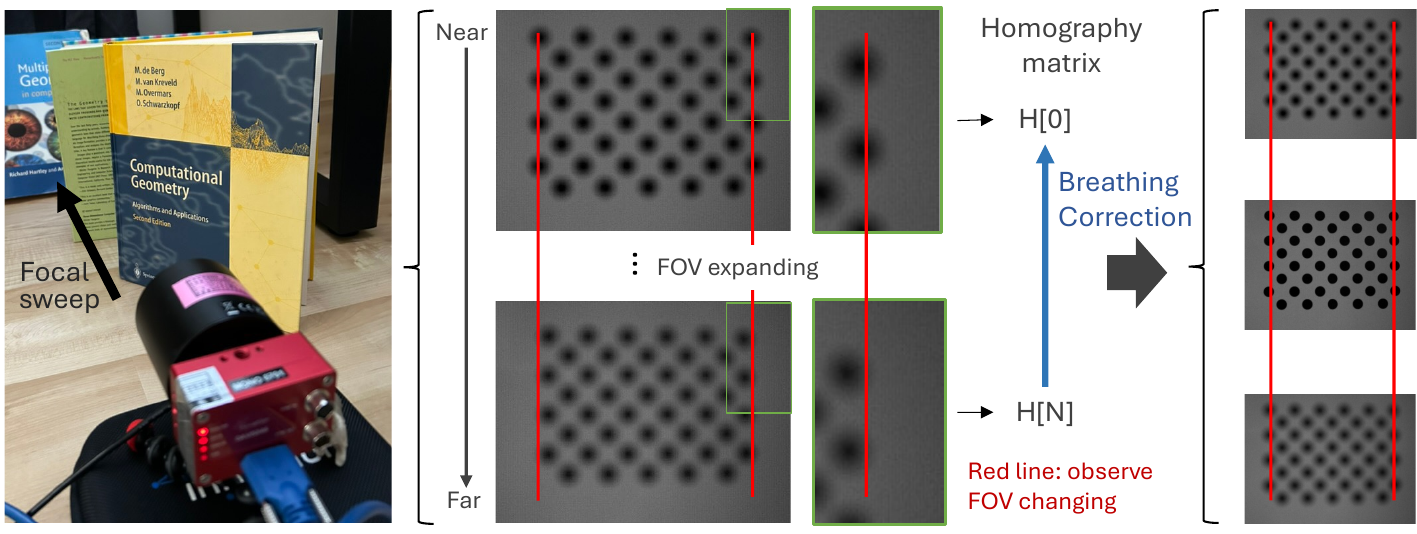}
    \caption{\textbf{Collecting real-captured data.}
    We captured the focal sweep events by event camera and computer-controlled lens.
    To avoid the impact of breathing, we use homography matrices $H[k] = \mathbb{R}^{3 \times 3}$ calculated by 330 images of a circular checkerboard for the correction.}
    \vspace{-5mm}
    \label{fig:real_capture}
    \end{center}
\end{figure*}
\setlength\textfloatsep{0pt}

\section{Proposed Method}
\label{sec:method}

\figurename~\ref{fig:flow} shows our framework. 
Given a sequence of event points ${\bm{e}_k=(x_i, y_i, t_i, p_i)}_{i \in [0, N-1]}$ that includes $N$ events, our framework obtains event focal stack, $\mathbf{V} \in \mathbb{R}^{W \times H \times B}$ by segmenting $\bm{e}_k$ into $B$ time bins of event voxel grids composed of the event frames of size $W$ (width) by $H$ (height). Here, $x, y$ indicates 2D location, $t$ and $p$ represent the timestamp when the event occurred and the polarity information indicating whether the brightness increased or decreased, respectively. 
Event focal stack $\mathbf{V}$ is fed into the depth map generation network to generate depth map $D_\text{pred} \in \mathbb{R}^{W \times H}$, which is our output.

For training, our method utilizes RGB image focal stacks from a simulation of the imaging process in virtual scenes using 3D computer graphics software, followed by event generation via an event simulator.
We used the synthetic data to leverage ground truth depth maps.
Note that image focal stacks are generated for training. {\bf{Our method relies solely on events during inference}}. For testing, we generate depth maps using the synthetic events and the events captured by a real event camera with sweeping lens.

The following subsections describe events from focus sweep to event focal stack in Sec.~\ref{sec:voxelize}, network design and loss in Sec.~\ref{sec:network} and fine-tuning and lens breathing correction in Sec.~\ref{sec:finetune}.

\subsection{Constructing Event Focal Stack}
\label{sec:voxelize}
Event points $\bm{e}$ form a 3D point cloud with high memory demands. We generate an event focal stack to reduce computational complexity by quantizing the event points along the time axis. 
One of the straightforward ways is to segment the event points into fixed time windows and record the number of events occurred at each pixels within that window on a single frame. However, this approach results in the loss of a significant amount of temporal information. Inspired by the task of generating gray-scaled images from event data~\cite{reconstruction}, we generate time-weighted event voxels~\cite{alex2018eventvoxel} to address this issue.

To generate $\mathbf{V}$ of $B \ll N$ from the observed events $\mathbf{e}_i$, we followed the voxelization method of Alex \etal ~\cite{alex2018eventvoxel}.
This `voxelization' process is outlined as follows:
\begin{enumerate}
    \item Normalize timestamps into the size, $B$ $(0 \leq t'_i < B)$.
    \item A polarity value, $p_i$, is linearly weighted depending on the distance to the two closest bins, $(b_k, b_{k+1})$.
    \item Store the values to the corresponding voxel location, $(x_i, y_i, b_k)$ and $(x_i, y_i, b_{k+1})$
    \item After collecting all events, voxel data is normalized by the min--max values in each dimension.
\end{enumerate}
We found the practical best bin size is $B=5$ (Tables \ref{tab:bin_compare_esim} and \ref{tab:bin_compare_DVS}), consistent with the work for a similar task \cite{reconstruction}.

\begin{figure*}[tb]
    \setlength\abovecaptionskip{0pt}
    \begin{center}
    \includegraphics[width=0.8\hsize]{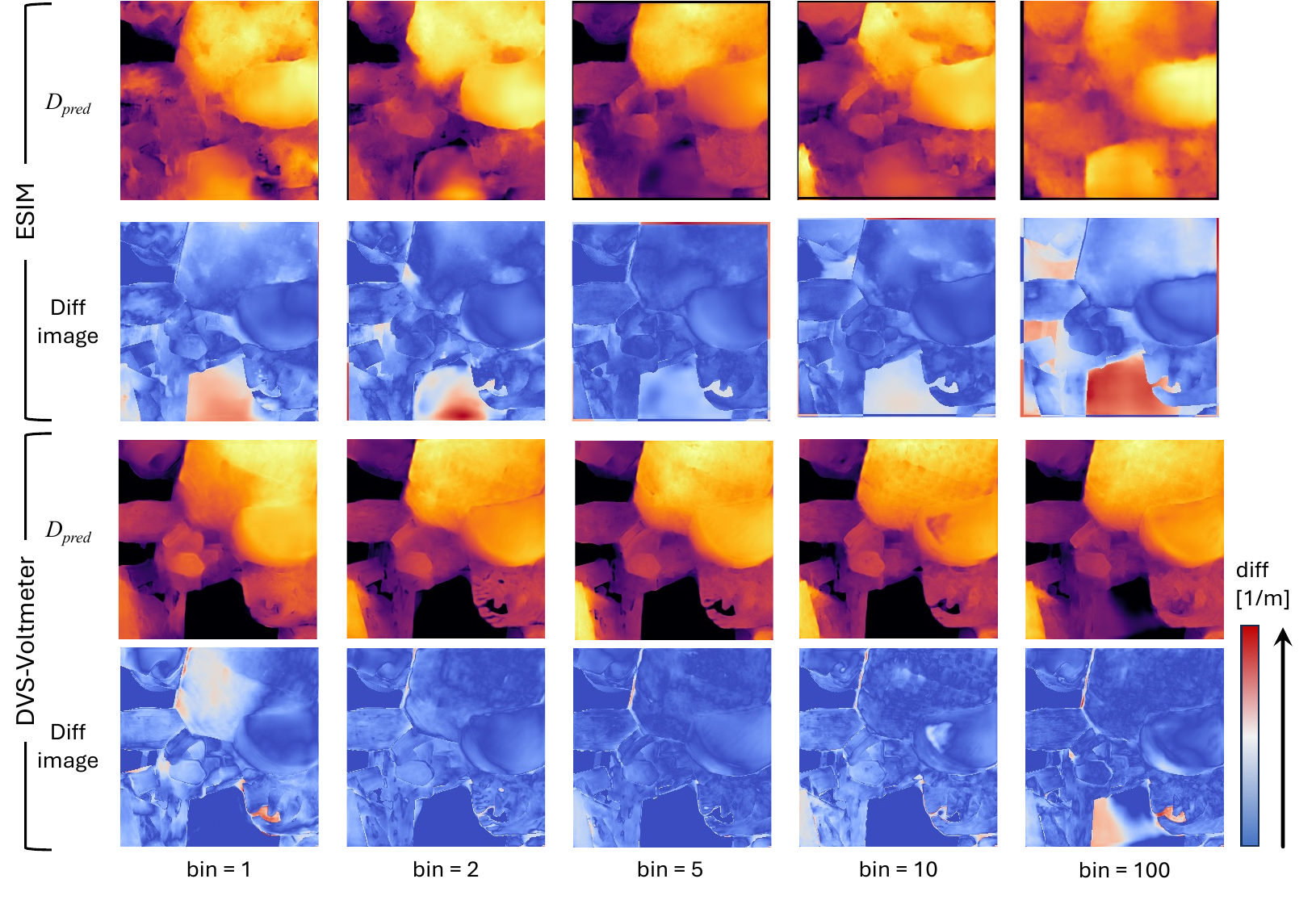}
    \caption{\textbf{Qualitative comparison of $bin$ impacts.}
    Inverse depth images indicate that the distance increases as the color transitions from orange to purple. 
    Differential images show that the error increases as the color transitions from blue to red.
    Both ESIM and DVS-Voltmeter, $bin=5$ shows the smallest error.}
    \vspace{-3mm}
    \label{fig:bin_compare}
    \end{center}
\end{figure*}
\setlength\textfloatsep{0pt}

\subsection{Depth Map Generation Network}\label{sec:network}
Following existing gray scale video reconstruction approach~\cite{reconstruction}, we use U-Net~\cite{unet} like architecture for a dense depth map estimation. To apply~\cite{reconstruction} to our task, we remove the recurrent functionality and modified the model to generate a single dense depth map.
Here, we treat inverse depth image as the output, in order to enhance the resolution of the near-field depth.

The input of the network is an event focal stack, which consists of $B=5$ event frames (each frame has a size of $256 \times 256$).
It is fed into the encoder, which has four 2D convolutional layers, to the feature map with a size of $512 \times 16 \times 16$.
Then, it is fed into an intermediate layer containing two convolutional layers, where it is transformed into $512 \times 16 \times 16$ feature map, and then passed through a four-layer decoder.
In each layer of the decoder, by adding the input features to the corresponding encoder features through skip connections, the location information of the events is conveyed.
Finally, the output layer converts the $32 \times 256 \times 256$ feature maps into $1 \times 256 \times 256$ single-channel gray-scale depth image.

With the variable $\theta$ that contains all trainable parameters, the training objective uses Mean Squared Error~(MSE) loss 
\begin{equation}
\mathcal{L}(\theta) = \frac{1}{WH}\sum^H_{x=1}\sum^W_{y=1}(D_\mathrm{gt}(x, y)-D_\mathrm{pred}(x, y))^2.
\label{mse_loss}
\end{equation}
For real-world data, we utilize pseudo $D_\mathrm{gt}$ generated by Depth Anything~\cite{depth_anything}.

\subsection{Lens Breathing Correction}
\label{sec:finetune}
To develop a method that works well not only on synthetic data but also in real scenes, we 
train the network on a large synthetic dataset and fine-tuning it with a small amount of real data. Unlike synthetic data with perfect optics, real event data is affected by slight distortions due to lens breathing during the focus sweep (\figurename~\ref{fig:real_capture}). Lens breathing is a phenomenon that involves varying fields of view as the lens focus distance changes. It is challenging to mitigate this effect optically for physical cameras. These distortions can be a major factor in the domain gap between synthetic and real data. Simply fine-tuning the model with raw real-captured data is insufficient to cope with our method effectively.

We geometrically correct lens breathing. Specifically, we calculate homography matrices $H[k] \in \mathbb{R}^{3 \times 3}$ from a reference plane in focus to the others at different focuses.
For accurate homography calculations, we use a circular checkerboard, which is more robust against lens defocus blur than a typical square grid pattern.
We recorded 330 images of a defocus-robust circular checkerboard at fixed distances of up to two meters.
Using the temporally closest homography matrix, we warped event coordinates, $(x_i, y_i)$.

\section{Experimental Settings}

\subsection{Datasets}\label{sec:dataset}
We create our synthetic and real datasets since no dataset is publicly available for this task.
Our strategy is to configure the best-performing network and voxelization approach and then evaluate the applicability to the real dataset.

\begin{table}[t]
    \centering
      \caption{Impact of the number of bins of ESIM.
      \textbf{Bold} and \underline{underline} fonts for the best and the second best scores.}
      \vspace{-3mm}
        \small
        \begin{tabular}[t]{rcc}
        \toprule
        Bin size $(B)$ & MAE [1/m] ($\downarrow$) & RMSE [1/m] ($\downarrow$)\\
        \midrule
        1 & 0.1536 & 0.1977 \\
        2 & \underline{0.1362}& \underline{0.1789} \\
        5 & \textbf{0.1335} & \textbf{0.1777} \\ 
        10 & 0.1401 & 0.1844 \\
        100 & 0.1878 & 0.2408 \\
        \bottomrule
        \label{tab:bin_compare_esim}
        \end{tabular}
\vspace{-7mm}
\end{table}

\begin{table}[t]
    \centering
      \caption{Impact of the number of bins of DVS-Voltmeter.
      \textbf{Bold} and \underline{underline} fonts for the best and the second best scores.}
      \vspace{-3mm}
        \small
        \begin{tabular}[t]{rcc}
        \toprule
        Bin size $(B)$ & MAE [1/m] ($\downarrow$) & RMSE [1/m] ($\downarrow$)\\
        \midrule
        1 & 0.1165 & 0.1519 \\
        2 & 0.0803 & 0.1065 \\
        5 & \underline{0.0762} & \underline{0.1022} \\ 
        10 & \textbf{0.0760} & \textbf{0.1006} \\
        100 & 0.0777 & 0.1045 \\
        \bottomrule
        \label{tab:bin_compare_DVS}
        \end{tabular}
\vspace{0mm}
\end{table}

\noindent
{\bf{Synthetic Dataset.}} 
To generate synthetic data for a wide variety of scenes, we rendered scenes of Thingi10K objects \cite{thingi10k} with random positions and materials using Blender \cite{blender}.
We set the camera parameters in Blender based on the lens parameters used in the real-world environment, with a focal length of 16mm, sensor size of 4.81mm (height) by 6.4mm (width), and an F-number of 2.3. However, it should be noted that these parameters in Blender do not faithfully replicate the actual camera.
After rendering image focal stacks of 600 focus distances, we simulated event data using representative simulators, ESIM \cite{simulator_esim} and DVS-Voltmeter \cite{simulator_dvs}, from the focal stack. The focuses were linearly placed at 0.2 to 2 meters in depth.
We also rendered corresponding depth images as the ground truth.
We eventually collected 1,000 scenes for training and 20 scenes for evaluations.

\begin{table}[t]
    \centering
      \caption{Impact of different polarity integration strategies of ESIM.
      \textbf{Bold} fonts for the best scores.}
        \label{tab:pn_compare_esim}
      \vspace{-3mm}
        \small
        \begin{tabular}[t]{rcc}
        \toprule
        Method & MAE [1/m] ($\downarrow$) & RMSE [1/m] ($\downarrow$)\\
        \midrule
        normal & 0.1335 & 0.1777 \\
        ppnn & \textbf{0.1256} & \textbf{0.1682} \\
        pnpn & 0.1312 & 0.1775 \\
        \bottomrule
        \end{tabular}
\vspace{-1mm}
\end{table}

\begin{table}[t]
    \centering
      \caption{Impact of different polarity integration strategies of DVS-Voltmeter.
      \textbf{Bold} fonts for the best scores.}
        \label{tab:pn_compare_DVS}
      \vspace{-3mm}
        \small
        \begin{tabular}[t]{rcc}
        \toprule
        Method & MAE [1/m] ($\downarrow$) & RMSE [1/m] ($\downarrow$)\\
        \midrule
        normal & \textbf{0.0762} & \textbf{0.1022} \\
        ppnn & 0.0838 & 0.1140 \\
        pnpn & 0.0881 & 0.1201 \\
        \bottomrule
        \end{tabular}
\vspace{3mm}
\end{table}

\noindent
{\bf{Event Simulator.}}
Obtaining a large amount of labeled data for event-based machine learning is challenging, especially for real-world datasets.
A potential solution is to create synthetic event data via an event simulator \cite{simulator_esim, simulator_dvs}.

An event simulator can virtually generate events by taking multiple luminance images as input.
ESIM \cite{simulator_esim} simulates event cameras by focusing on the visual signal and adapting the frame sampling rate according to scene dynamics.
DVS-Voltmeter \cite{simulator_dvs} simulates events based on the physical properties of dynamic vision sensors (DVS) such as voltage changes.

Despite these attempts, we have observed critical gaps between real and synthetic events in our real-world scenario.
We resolve this problem by exploring the best-practical event focal stack configuration on a large-scale synthetic dataset and fine-tuning the pre-trained network with fewer real-scene datasets.

\noindent
{\bf{Real Dataset.}} 
To further showcase our method’s applicability, we also tested our method with real captured event data. We used DAVIS346 equipped with a computer-controlled lens (LensConnect DL1623UC-MPY) to capture event data synchronized with gray scale images. 
One of the advantages of our task—depth map estimation using events—is its robustness in low-light conditions. Therefore, we collected data from the same scene under two different lighting environments: a well-lit scene and a low-light environment. In total, we recorded data from 60 scenes (30 scenes $\times$ 2 lighting conditions).

Recording ground truth depth maps for these real scenes was challenging. Therefore, we used the state-of-the-art depth estimator, Depth Anything~\cite{depth_anything} for luminance images, as an alternative ground truth.
Since Depth Anything does not perform well under low-light conditions, we used images under well-lit conditions for low-light conditions. 
We used 50 scenes, including both well-lit and low-light conditions, for fine-tuning when applying real data for inference. The remaining 10 scenes used for inference were captured with depth distributions and objects different from those used for fine-tuning.

\subsection{Baseline Method}\label{sec:baseline}
We compare our approach with a depth-from-defocus method for the color domain, Focus on Defocus \cite{image_focal_depth_estimation}.
For a fair comparison, we trained the Focus on Defocus network on our synthetic dataset and used the same bin size $(B=5)$.
Although this network accepts three-channel color images, our real dataset consists of only gray-scale images.
For better applicability to real scenes, we converted our original color images into three-channel gray-scale images and trained the network.

\subsection{Evaluation Metrics}\label{sec:measure}
We calculated mean absolute error (MAE) $[1/m]$ $(\downarrow)$, and root mean squared error (RMSE) $[1/m]$ $(\downarrow)$, for quantitative metrics:
\[ \displaystyle 
    MAE = \sum_W \sum_H \left|D_\text{pred} - D_\text{gt}\right| / (WH)
\]
\[ \displaystyle 
    RMSE = \sqrt{\sum_W \sum_H (D_\text{pred} - D_\text{gt})^2 / (WH)}.
\]
MAE and RMSE quantify the average magnitude of errors between prediction and ground truth, and RMSE is more sensitive to outliers.
We also discuss qualitative results.

\subsection{Implementation Details}
For both the baseline and our methods, we trained for 200 epochs using the Adam optimizer~\cite{adam} with a learning rate of $1.0\times10^{-4}$.
We randomly rotated the data in $W \times H$ dimensions for data augmentation.
We performed the same evaluations 10 times to mean out the influence of the random seeds for training.

\subsection{Parameter Validations}\label{sec:preliminary}
There are multiple ways to determine the event focal stack (such as differences in bin size and event polarity). Furthermore, by comparing the events generated by event simulators (ESIM, DVS-Voltmeter) with real events, we aim to select and evaluate suitable simulator for this task.

\noindent
{\bf{Bin size.}}
The bin size, $B$, alters the temporal resolution of events in an event focal stack, $\mathbf{V}$, which can affect the overall performance.
Although the prior work \cite{reconstruction} found $B=5$ for their empirical best, the parameter can be task-dependent.
Therefore, using our synthetic dataset, we evaluated our network with differently sized bins of event focal stacks and their impacts on performance.
We measured the errors under different bin sizes of $\{1, 2, 5, 10, 100\}$.

Tables~\ref{tab:bin_compare_esim} and \ref{tab:bin_compare_DVS} show the highest performance on $B=5$ with the events by ESIM and $B=10$ with the events by DVS-Voltmeter, respectively.
However, \figurename~\ref{fig:bin_compare} shows the plausibly lower errors with $B=5$ for both event simulators.
Considering that the second-highest score for DVS-Voltmeter is $B=5$, which is more compact and consistent with the prior finding \cite{reconstruction}, we conclude to take $B=5$.

\begin{figure}[tb]
    \setlength\abovecaptionskip{0pt}
    \begin{center}
    \includegraphics[width=1.0\hsize]{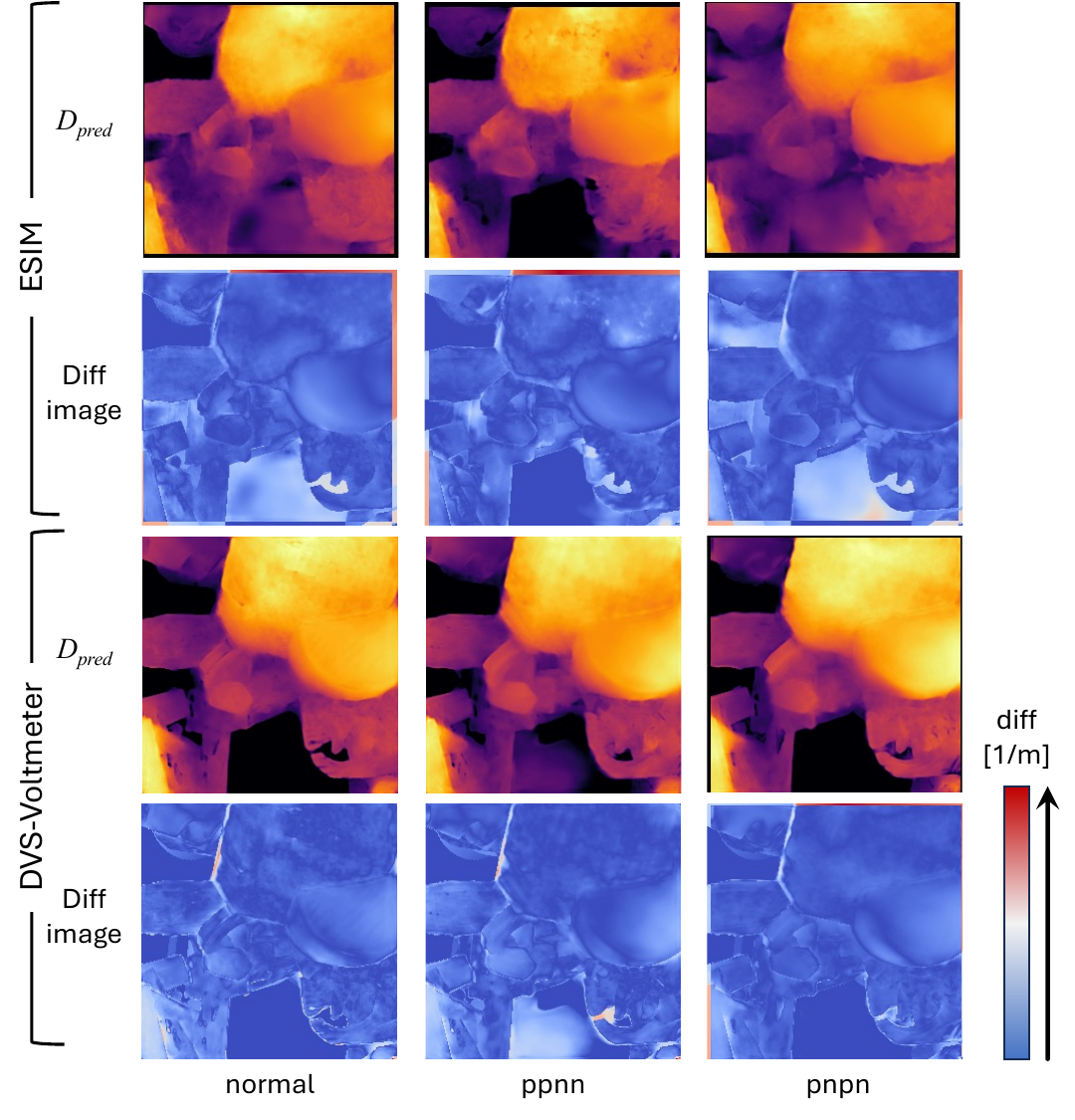}
    \caption{\textbf{Qualitative comparison of the impact of polarity integration.}
    It seems no significant differences between $normal$, $ppnn$ and $pnpn$.}
    \label{fig:pn_compare}
    \end{center}
\end{figure}
\setlength\textfloatsep{10pt}

\noindent
{\bf{Polarity integration.}}
Another design choice in the voxelization is the arrangement of negative and positive polarities.
In the voxelization (\textit{normal}) in Section~\ref{sec:method}, positive and negative polarity events that fall into a pixel mean out.
For smaller $B$, this can happen more often.
To mitigate this, one may separately store the positive (\textit{p}) and negative (\textit{n}) responses in different channels (i.e., 10 channels for $B=5$).
The \textit{pn} order would matter for this option (i.e., \textit{pnpn} and \textit{ppnn}).
Tables \ref{tab:pn_compare_esim} and \ref{tab:pn_compare_DVS} indicate that 
the different polarity arrangements do not have a significant impact.
Therefore, we take the \textit{normal} approach.

\begin{figure}[tb]
\setlength\abovecaptionskip{0pt}
    \begin{center}
    \includegraphics[width=1.0\hsize]{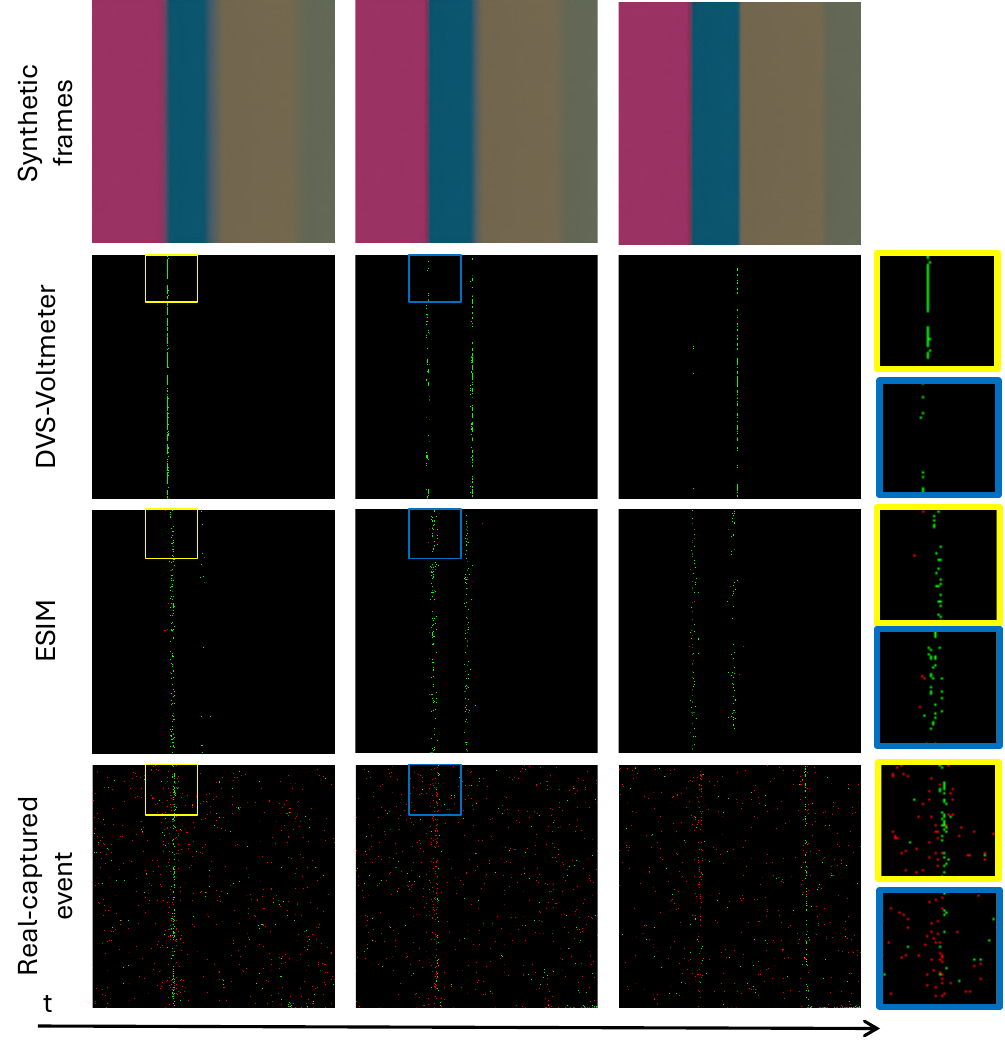}
    \caption{\textbf{Qualitative comparison of events using ESIM, DVS-Voltmeter and real-captured event.} The captured scene is one in which boxes are arranged to become progressively more distant from left to right. 
    Although the real-captured events can be observed negative events (red dots) and noisy events, the events generated by ESIM and DVS-Voltmeter are hard to observe negative events and appear less noisy.
    }
    \vspace{2mm}
    \label{fig:ESIM_DVS}
    \end{center}
\end{figure}
\setlength\textfloatsep{0pt}

\begin{table}[t]
    \centering
      \caption{Comparison on the synthetic dataset.
      \textbf{Bold} and \underline{underline} fonts for the best and the second best scores.
      }
      \vspace{-3mm}
        \small
        \begin{tabular}[t]{lcc}
        \toprule
        Method & MAE [1/m] & RMSE [1/m]\\
        & ($\downarrow$) & ($\downarrow$) \\
        \midrule
        Focus on Defocus \cite{image_focal_depth_estimation} & 0.1606 & 0.2027 \\
        Ours (ESIM) & \underline{0.1335} & \underline{0.1777} \\
        Ours (DVS-Voltmeter) & \textbf{0.0762} & \textbf{0.1022}\\
        \bottomrule
        \label{tab:compare_synth}
        \end{tabular}
\vspace{3mm}
\end{table}

\begin{figure*}[tb]
\setlength\abovecaptionskip{0pt}
\begin{center}
\includegraphics[width=0.95\hsize]{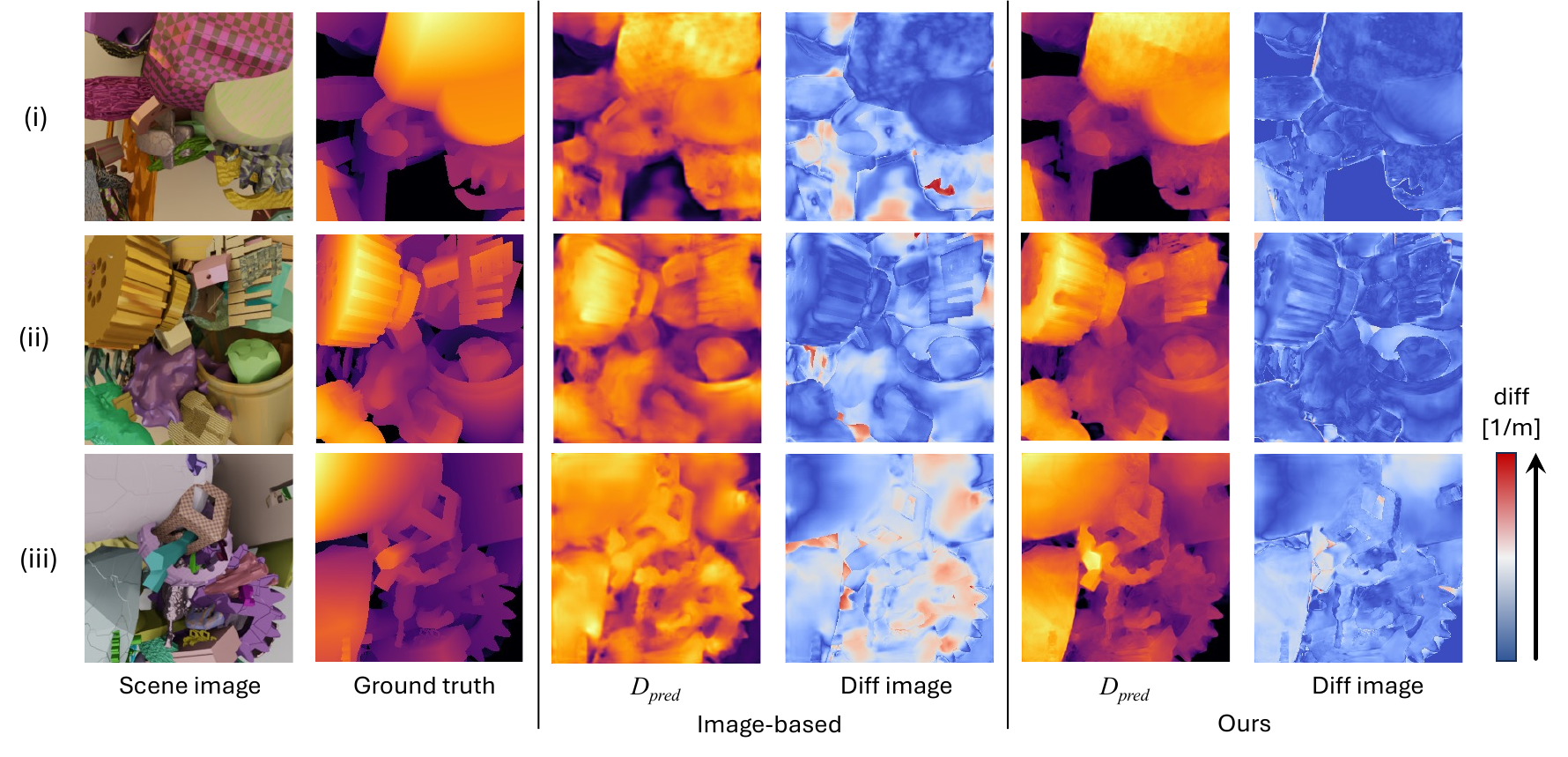}
\caption{\textbf{Qualitative comparison using synthetic data generated by DVS-Voltmeter.}
Ours have a small margin of error, and the depth images do not become blurred.}
\label{fig:compare_synth}
\end{center}
\end{figure*}
\setlength\textfloatsep{0pt}
\setlength\intextsep{0pt}

\noindent
{\bf{Event Simulator.}}
The reproducibility of event simulators matters for our task.
We prepared a physical scene and reproduced it in Blender to compare real and simulated events under a shared condition.
\figurename~\ref{fig:ESIM_DVS} shows the results of events from focal sweep.
The scenes consist of color boxes placed at progressively farther distances.
The scene depths get further from the left to the right in the image space.

During a focal sweep, positive events appear when lights converge to a point in focus, and then negative events appear due to the diffusion of lights.
The patterns would vary depending on the luminance at pixel locations and those in adjacent pixel locations.

DVS-Voltmeter shows events drawing unnaturally sharp lines and fails to reproduce reasonable noises.
We consider that the reason for the noiseless results is that the synthetic scene has no noise, but the model is built upon real images, which inherently contain noise.
ESIM shows scattered and noisy events that look closer to their real counterparts. To obtain this appearance, we needed to set the event threshold parameter to 0.08.
Nevertheless, both types of events do not show many negative events (in red), which suggests the limited feasibility of the simulators in our task. This led us to the fine-tuning approach for the real dataset.

\section{Experiments and Discussion}
\label{sec:discussion}

\paragraph{Synthetic dataset.}
\tablename~\ref{tab:compare_synth} summarizes the quantitative results. 
Our method shows superior results to our baseline, Focus on Defocus~\cite{image_focal_depth_estimation}, regardless of the event simulators.
Ours with ESIM and DVS-Voltmeter reduced the error by approximately 1.2 and 2.0 times compared to the baseline, respectively.
With DVS-Voltmeter, ours performed the best.

\figurename~\ref{fig:compare_synth} shows estimated depth maps and corresponding difference images of the baseline and ours with DVS-Voltmeter in three example scenes.
Contrary to the baseline, which loses sharp edges and smooth surfaces, ours keeps the scene structures and is less affected by the texture details.
The baseline shows high errors over the image space, and ours successfully suppresses them.

The results indicate that our event focal stack can preserve denser edge information in a smoother focal sweep, and our simple network can interpret the captured information into a depth map.
On the other hand, an image focal stack has rather sparser or discretized information than ours.

\begin{figure*}[tb]
\setlength\abovecaptionskip{0pt}
    \begin{center}
    \includegraphics[width=1.0\hsize]{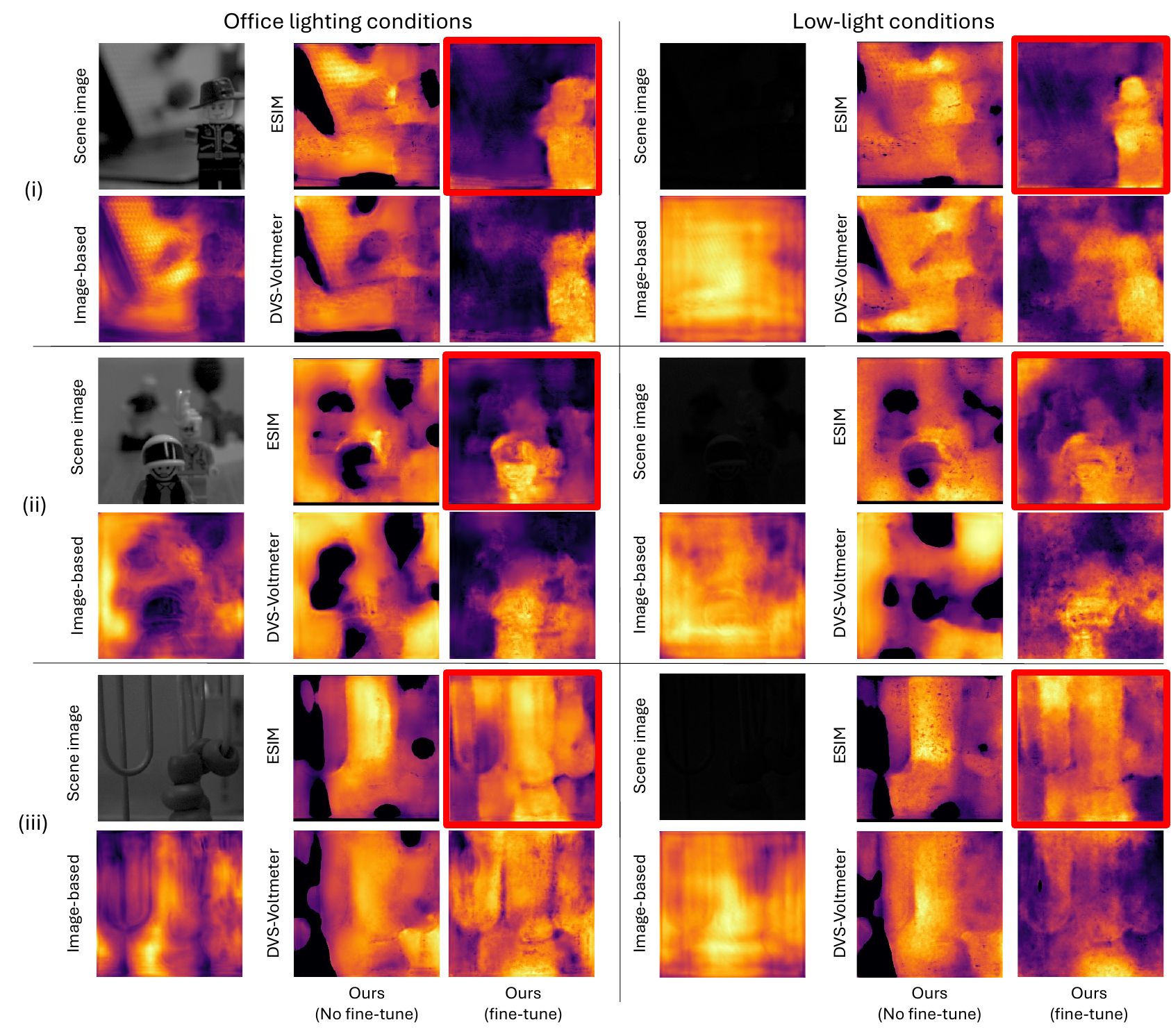}
    \caption{\textbf{Qualitative comparison using real-captured data at the condition of two light environment.}
    The rows show the same scene.
    We prepared models trained on synthetic events generated by ESIM and DVS-Voltmeter, and compared the results with and without fine-tuning on real data.
    The results highlighted with red frames are the most feasible results with clear shapes and spatial relationships.
    }
    \vspace{-7mm}
    \label{fig:compare_real}
    \end{center}
\end{figure*}
\setlength\textfloatsep{0pt}

\paragraph{Real dataset.}
\figurename~\ref{fig:compare_real} (left) shows depth maps of the baseline and ours in three real scenes under office lighting conditions.
The baseline method and ours without the real-scene fine-tuning tend to show front-back reversed depths.
Ours even shows some holes.
With fine-tuning, ours can mitigate both the reversed depth and holes and present reasonable overall depths, overcoming the domain gap between real and simulated events.
However, details such as thin structures and edges are more ambiguous than results in the synthetic dataset.
From these results, it can be inferred that ours with fine-tuning is superior to image-based method and there is a domain gap between the synthetic event data used for training and the real-captured event data.
Also, they shows fine-tuning with real event is effective method to overcome this problem.
Compared to ours with ESIM and DVS-Voltmeter, ESIM can present sharper and smoother scene structures and appears to be a superior choice.
This draws the same conclusion as in Section \ref{sec:preliminary}.

\figurename~\ref{fig:compare_real} (right) demonstrates the robustness of our approach under extreme conditions (i.e., low-light conditions).
Since intensity images capture almost no information, the baseline method fails to estimate any reasonable depth maps.
Ours relies on events sensitive to subtle changes in the imaging sensor and can still grasp the depth information.
Same as in the office lighting conditions, without fine-tuning, ours shows holes and inconsistent depths.

\paragraph{Limitation.}
Our method shows inferior results on textureless surfaces since it lacks events of depth clues.
Also, our method is designed for static scenes. If dynamic content appears, events on the content must be spatially aligned.
Furthermore, the performance of our imaging depends on the focus sweep mechanisms. While we used a mechanical lens, a focus-tunable lens can provide a faster focus sweep.

The susceptibility of image- and event-based methods in different environments can vary. Different image blur shapes can have a negligible impact on the former because defocus blurs spread evenly. Event-based methods focusing on brightness changes are more affected by the position and timing of events from the blur. This makes event-based methods more sensitive to variations in camera parameters, increasing the significance of domain gaps.

\section{Conclusion}
In this paper, towards robust method even under low-light conditions, we proposed dense depth estimation from an event focal stack.
To this end, we proposed a framework and training strategies, including fine-tuning for real-scene datasets.
We validated event focal stack data structure and configurations for the best performance.
We compared ours with the baseline method using an image focal stack for dense depth map estimation.
We identified the domain gap between real and virtual events from lens focus sweeping in our task.
Although the image blurs by defocus have fewer domain shifts between real and virtual worlds, events from such blurs are strongly dependent on event simulators.
The quantitative and qualitative results showcased that ours, after real-dataset fine-tuning, is more robust.
Our future work includes improving algorithms and simulators to further reduce the domain gap between synthetic and real-world data.

\noindent
{\textbf{{Acknowledgement}}
This work was partially supported by
JST Presto JPMJPR22C1,
Keio University Academic Development Funds,
the Austrian Science Fund FWF (grant no. P33634), and JSPS KAKENHI JP23H03422.

{\small
\bibliographystyle{ieee_fullname}
\bibliography{egbib}

\begin{thebibliography}{10}\itemsep=-1pt

\bibitem{stereo_depth}
{Alex Kendall, Hayk Martirosyan, Saumitro Dasgupta, Peter Henry, Ryan Kennedy, Abraham Bachrach, Adam Bry}.
\newblock End-to-end learning of geometry and context for deep stereo regression.
\newblock In {\em ICCV}, pages 66--75, 2017.

\bibitem{depth_from_defocus}
{Alex Paul Pentland}.
\newblock A new sense for depth.
\newblock {\em IEEE TPAMI}, PAMI-9(4):523--531, 1987.

\bibitem{blender}
{Blender Foundation}.
\newblock The {Blender} project - free and open 3d creation software, Accesed: 2023.

\bibitem{jiang2024learning}
Jiang Chenxu, Lin Mingyuan, Zhang Chi, Wang Zhenghai, and Yu Lei.
\newblock Learning monocular depth from focus with event focal stack.
\newblock {\em arXiv preprint arXiv:2405.06944}, 2024.

\bibitem{first_depth}
{D. Eigen and R. Fergus}.
\newblock Predicting depth, surface normals and semantic labels with a common multi-scale convolutional architecture.
\newblock In {\em ICCV}, pages 2650--2658, 2015.

\bibitem{adam}
{Diederik P. Kingma and Jimmy Ba}.
\newblock Adam: A method for stochastic optimization.
\newblock {\em ICLR}, 2014.

\bibitem{mixed_reality}
Ruofei Du, Eric~Lee Turner, Maksym Dzitsiuk, Luca Prasso, Ivo Duarte, Jason Dourgarian, Joao Afonso, Jose Pascoal, Josh Gladstone, Nuno~Moura e Silva~Cruces, Shahram Izadi, Adarsh Kowdle, Konstantine Nicholas~John Tsotsos, and David Kim.
\newblock Depthlab: Real-time 3d interaction with depth maps for mobile augmented reality.
\newblock In {\em Proceedings of the 33rd Annual ACM Symposium on User Interface Software and Technology}, pages 829--843, 2020.

\bibitem{deep_depth_from_focus}
{Fengting Yang, Xiaolei Huang, and Zihan Zhou}.
\newblock Deep depth from focus with differential focus volume.
\newblock In {\em CVPR}, pages 12632--12641, 2022.

\bibitem{event_camera}
{Guillermo Gallego, Tobi Delbru\"{u}ck, Garrick Orchard, Chiara Bartolozzi, Brian Taba, Andrea Censi, Stefan Leutenegger, Andrew J. Davison, J\"{o}rg Conradt, Kostas Daniilidis, Davide Scaramuzza}.
\newblock Event-based vision: A survey.
\newblock {\em IEEE TPAMI}, 44(1):154--180, 2019.

\bibitem{haessig2019spiking}
Germain Haessig, Xavier Berthelon, Sio-Hoi Ieng, and Ryad Benosman.
\newblock A spiking neural network model of depth from defocus for event-based neuromorphic vision.
\newblock {\em Scientific reports}, 9(1):3744, 2019.

\bibitem{event_all_in_focus}
{Hanyue Lou, Minggui Teng, Yixin Yang, and Boxin Shi}.
\newblock All-in-focus imaging from event focal stack.
\newblock In {\em CVPR}, pages 17366--17375, 2023.

\bibitem{reconstruction}
{Henri Rebecq, Ren\'{e} Ranftl, Vladlen Koltun, and Davide Scaramuzza}.
\newblock High speed and high dynamic range video with an event camera.
\newblock {\em IEEE TPAMI}, 43(6):1964--1980, 2019.

\bibitem{acoustic_depth}
Go Irie, Takashi Shibata, and Akisato Kimura.
\newblock Co-attention-guided bilinear model for echo-based depth estimation.
\newblock In {\em ICASSP}, pages 4648--4652, 2022.

\bibitem{Ishikawa2023FS2MPI}
Reina Ishikawa, Hideo Saito, Denis Kalkofen, and Shohei Mori.
\newblock Multi-layer scene representation from composed focal stacks.
\newblock {\em IEEE TVCG}, 29(11):4719--4729, 2023.

\bibitem{e2depth}
{Javier Hidalgo-Carri\'{o}, Daniel Gehrig and Davide Scaramuzza}.
\newblock Learning monocular dense depth from events.
\newblock In {\em 3DV}, pages 534--542, 2020.

\bibitem{image_focal_depth_estimation}
{Maxim Maximov, Kevin Galim, and Laura Leal-Taixe}.
\newblock Focus on defocus: Bridging the synthetic to real domain gap for depth estimation.
\newblock In {\em CVPR}, pages 1071--1080, 2020.

\bibitem{m4depth}
{Micha\"{e}l Fonder, Damien Ernst, Marc Van Droogenbroeck}.
\newblock M4depth: Monocular depth estimation for autonomous vehicles in unseen environments.
\newblock {\em Sensors}, 22(23):1--22, 2022.

\bibitem{unet}
{Olaf Ronneberger, Philipp Fischer, and Thomas Brox}.
\newblock U-net: Convolutional networks for biomedical image segmentation.
\newblock In {\em International Conference on Medical image computing and computer-assisted intervention}, pages 234--241, 2015.

\bibitem{thingi10k}
{Q. Zhou and A. Jacobson}.
\newblock Thingi10k: A dataset of 10,000 3d-printing models.
\newblock {\em arXiv preprint}, 2016.

\bibitem{simulator_esim}
Henri Rebecq, Daniel Gehrig, and Davide Scaramuzza.
\newblock {ESIM}: an open event camera simulator.
\newblock {\em Conf. on Robotics Learning (CoRL)}, Oct. 2018.

\bibitem{midas}
{Ren\'{e} Ranftl, Katrin Lasinger, David Hafner, Konrad Schindler, and Vladlen Koltun}.
\newblock Towards robust monocular depth estimation: Mixing datasets for zero-shot cross-dataset transfer.
\newblock {\em IEEE TPAMI}, 44(10):1623--1637, 2022.

\bibitem{simulator_dvs}
{Songnan Lin, Ye Ma, Zhenhua Guo, and Bihan Wen}.
\newblock Dvs-voltmeter: Stochastic process-based event simulator for dynamic vision sensors.
\newblock In {\em ECCV}, pages 578--593, 2022.

\bibitem{cv_book}
Richard Szeliski.
\newblock {\em Computer Vision: Algorithms and Applications}.
\newblock Springer Nature, 2022.

\bibitem{event_depth_EReformer}
{Xu Liu, Jianing Li, Xiaopeng Fan and Yonghong Tian}.
\newblock Event-based monocular dense depth estimation with recurrent transformers.
\newblock {\em arXiv preprint}, 2022.

\bibitem{depth_anything}
Lihe Yang, Bingyi Kang, Zilong Huang, Xiaogang Xu, Jiashi Feng, and Hengshuang Zhao.
\newblock Depth anything: Unleashing the power of large-scale unlabeled data.
\newblock In {\em CVPR}, pages 10371--10381, 2024.

\bibitem{motion_blur_depth}
{Zhen, Ruiwen and Stevenson, Robert L.}
\newblock Motion deblurring and depth estimation from multiple images.
\newblock In {\em ICIP}, pages 2688--2692, 2016.

\bibitem{alex2018eventvoxel}
Alex~Zihao Zhu, Liangzhe Yuan, Keneth Chaney, and Kostas Daniilidis.
\newblock Unsupervised event-based optical flow using motion compensation.
\newblock In {\em ECCVW}, 2018.

\end{thebibliography}
}

\end{document}